%% file: 5018.tex
\documentclass[runningheads]{llncs}
\usepackage{graphicx}
\usepackage{orcidlink}
\usepackage{tikz}
\usepackage{comment}
\usepackage{amsmath,amssymb} % define this before the line numbering.
\usepackage{color}
\usepackage{cite}
\usepackage{makecell}
\usepackage{multirow}
\usepackage{multicol}
\usepackage{bm}
\usepackage{amsmath,graphicx}
\usepackage{amssymb}
\usepackage{pifont}
\usepackage{booktabs}
\usepackage{lipsum}
\usepackage{arydshln}
\usepackage{array}
\newcolumntype{x}[1]{>{\centering\arraybackslash\hspace{0pt}}p{#1}}
\usepackage{wrapfig}
\usepackage{capt-of}
\newcommand{\ie}{\textit{i}.\textit{e}.}
\newcommand{\eg}{\textit{e}.\textit{g}.}

\newcommand{\cmark}{\ding{51}}
\newcommand{\xmark}{\ding{55}}
\newsavebox\tmpbox
\usepackage{hyperref}
\hypersetup{
	colorlinks=true,
	linkcolor=red,
	filecolor=magenta,      
	urlcolor=black,
}

\usepackage[capitalize]{cleveref}
\crefname{section}{Sec.}{Secs.}
\Crefname{section}{Section}{Sections}
\Crefname{table}{Table}{Tables}
\usepackage[accsupp]{axessibility}  % Improves PDF readability for those with disabilities.

\begin{document}
% \linenumbers
\pagestyle{headings}
\mainmatter
\def\ECCVSubNumber{5018}  % Insert your submission number here

\title{VisageSynTalk:\\Unseen Speaker Video-to-Speech Synthesis\\via Speech-Visage Feature Selection} % Replace with your title

\begin{comment}
\titlerunning{ECCV-22 submission ID \ECCVSubNumber} 
\authorrunning{ECCV-22 submission ID \ECCVSubNumber} 
\author{Anonymous ECCV submission}
\institute{Paper ID \ECCVSubNumber}
\end{comment}
%******************

%\begin{comment}
\titlerunning{VisageSynTalk: Unseen Speaker Video-to-Speech Synthesis}
\author{Joanna Hong\orcidlink{0000-0003-4182-1000}\and
	Minsu Kim\orcidlink{0000-0002-6514-0018}\and
	Yong Man Ro\orcidlink{0000-0001-5306-6853}\index{Ro, Yong Man}}
\authorrunning{J. Hong et al.}
%\vspace{-0.5cm}
\institute{
	{Image and Video Systems Lab, School of Electrical Engineering, KAIST, South Korea}\\
	\email{\{joanna2587, ms.k, ymro\}@kaist.ac.kr}}
%\end{comment}
%******************
\maketitle

%\vspace{-0.3cm}
\begin{abstract}
The goal of this work is to reconstruct speech from a silent talking face video. Recent studies have shown impressive performance on synthesizing speech from silent talking face videos. However, they have not explicitly considered on varying identity characteristics of different speakers, which place a challenge in the video-to-speech synthesis, and this becomes more critical in unseen-speaker settings. %Distinct from the previous methods,
Our approach is to separate the speech content and the visage-style from a given silent talking face video. By guiding the model to independently focus on modeling the two representations, we can obtain the speech of high intelligibility from the model even when the input video of an unseen subject is given. To this end, we introduce speech-visage selection that separates the speech content and the speaker identity from the visual features of the input video. The disentangled representations are jointly incorporated to synthesize speech through visage-style based synthesizer which generates speech by coating the visage-styles while maintaining the speech content.
% The proposed module outputs speech selective masks containing the distinctive scores of the speech content information in the visual features of a talking face video, while leaving out its speaker identity attributes. Whereas, the speaker identity attributes are obtained from the inverse of the speech selective masks. 
%With the above obtained speech content features and identity features, 
% We further introduce a visage-style based synthesizer which takes the content features as an input and sequentially coats the visage-styles of the extracted identity features to generate speech. %Thus, the proposed framework can separately focus on modeling the speech content and generating the speech with one's visage-style.
Thus, the proposed framework brings the advantage of synthesizing the speech containing the right content even with the silent talking face video of an unseen subject. We validate the effectiveness of the proposed framework on the GRID, TCD-TIMIT volunteer, and LRW datasets. %The synthesized speech can be heard in supplementary materials. % on both multi-speaker independent and dependent settings. 
%We also analyze the effectiveness of the multiple speech selective masks.
\end{abstract}

%\vspace{-0.5cm}
\section{Introduction}
\label{sec:intro}
%\vspace{-0.2cm}

Imagine a subway station packed with people, and a middle-aged woman next to you appears to ask you something. It is hard for you to understand her because of the noise of an incoming subway, %but you cannot hear well because of the noise of an incoming subway. 
so you try to follow her by looking at her face and mouth movements and infer what she tries to say. Then, you can finally understand and give an answer to her. These days, people frequently encounter these kinds of situations, not only in real-time but also in silent video conferences, corrupted video messages, and even conversations with a speech-impaired person \cite{burnham2013hearingeye}. In order to help these situations, there has been much research, namely lip-reading, on recognizing speech from silent or audio-corrupted videos. 

Video-to-speech synthesis is one of the lip-reading techniques, which reconstructs speech from silent talking face videos. It has the advantage of not requiring extra human annotations (\ie, text), while other conventional text-based lip-reading techniques need them \cite{stafylakis2017resnetlstm,assael2016lipnet}. Nevertheless, video-to-speech synthesis is considered as challenging since it is expected to represent not only the speech content but also the identity characteristics (\eg, voice) of the speaker. Thus, it is difficult to be applied in unseen, even multi-speaker, settings. %With the great development of deep learning, 
There has been remarkable progresses in video-to-speech synthesis \cite{ephrat2017improved, ephrat2017vid2speech, michelsanti2020vocoderbased,prajwal2020lip2wav,vougioukas2019ganbased,kim2021memory,hong2021speech}, especially with few speakers. While they have shown impressive performances, they have not explicitly considered the varying identity characteristics of different speakers, thus not investigated well in unseen multi-speaker setting.

%However, lip to speech synthesis has been considered as a challenging problem due to the ambiguity that homophenes intrinsically contain – different characters that produce the same lip movements, such as m, p, and b.

%Thanks to the great development of deep learning, there has been remarkable progress in the video-to-speech synthesis.\cite{ephrat2017improved, ephrat2017vid2speech, michelsanti2020vocoderbased} proposed to reconstruct auditory features from lip movements videos, \cite{prajwal2020lip2wav} proposed to utilize a sequence-to-sequence (Seq2Seq) architecture to capture the context, and \cite{vougioukas2019ganbased,mira2021miragan} proposed to directly synthesize the raw waveform of speech using a 1-dimensional Generative Adversarial Network (GAN). While these works have shown impressive performance, they have not considered the varying identity characteristics of different speakers, which places a challenge in the video-to-speech synthesis. This challenge becomes more critical when it comes to unseen-speaker settings or combined multi-speaker settings of speech synthesis, which are not addressed well in the previous works. 

To alleviate the challenge, we draw inspiration from human intuition in predicting a silent speech. When a silent talking video – seen or unseen – is given, humans firstly look at the entire appearance that represents the speaker’s character (e.g., gender and age) and then predict the speech sound based on the lip movements \cite{chen2001audiovisual}. By mimicking the human speech predicting process, we propose to learn to disentangle the lip movements (\ie, speech content) and the visage appearances (\ie, identities) from a silent talking face video and to predict the speech by jointly modeling the two disentangled representations. In doing so, it is promising that the model can reconstruct speech containing correct content from even unseen speaker's talking face videos.

%Early works \cite{ephrat2017improved, ephrat2017vid2speech} have incorporated convolution neural networks (CNNs) based encoder-decoder module to reconstruct acoustic speech signals from silent videos of a speaking person. A sequence-to-sequence network (Seq2Seq) \cite{prajwal2020lip2wav} has also been utilized predicting auditory features conditioned on both encoded visual contexts and previous predictions to reconstruct finer sounds. While promising performance of Seq2Seq architecture, it requires heavy training and inference time due to the sequential nature of the architecture. The recent works \cite{vougioukas2019ganbased,mira2021miragan} have brought generative adversarial network to synthesize a raw waveform from the video using 1-dimensional GAN. 

%they have faced the innate challenge of lip to speech synthesis that can be vulnerable to the varying identity characteristics of different speakers. 

%Thus, we come up with devising a network that can separate a speaker’s identity and the speech content, so that the network can distinguish one’s outward features and the speech from a silent talking face video.

%\vspace{-0.05cm}
In this paper, we introduce a novel framework for video-to-speech synthesis. It consists of speech-visage feature selection module that separates speech content and visage-style (\ie, identity) from a given talking face video. The proposed module exploits a deep learning-based feature selection \cite{gui2019afs,liao2021featureselectionusingbatchwiseattentuation} with feature transformation and normalization, which is jointly trained with the entire model in an end-to-end manner. The proposed module outputs speech selective masks, each of which contains the distinctive score of the speech content information in the visual feature of a talking face video while leaving out its speaker identity attributes. From the masks, the speech content features and the identity features can be separately driven. With the obtained two distinctive features through the speech-visage feature selection module, we introduce a visage-style based synthesizer, called VS-synthesizer. Inspired by \cite{karras2019stylegan, chen2021againvc} \cite{chen2021againvc}, the content features are taken into the VS-synthesizer as input, and the encoded content features are sequentially coated with the visage-styles of extracted identity features. 

%\vspace{-0.05cm}
In order to guide the proposed framework, two learning methods are proposed: visual- and audio-identification. In visual-identification learning, we guide the network to produce the same identity features when they are from the same subject and to predict right subject identity from the identity features. Through audio-identification learning, we expect that the network well predicts the correct subject identity from the generated mel-spectrogram, even when the different identity features are coated in the original speech content features.

%\vspace{-0.05cm}
Through the proposed framework, the model can separately focus on modeling the speech content and generating the speech with target speaker's appearance. It brings the advantage of synthesizing speech containing the right content even if a silent talking face video of an unseen subject is given. Moreover, the proposed framework can synthesize speech with different visage-styles while maintaining the original content. Our key contributions are as follows: (1) To the best of our knowledge, it is the first time to directly tackle the challenge induced from varying visage-styles of different speakers, in video-to-speech synthesis by separating the identity and speech content. (2) We design a speech-visage feature selection for masking identity attributes from a talking face video while maintaining speech content, and vice versa. (3) To guarantee the disentanglement of speech content from identity, we propose two learning methods: visual-identification and audio-identification. 
\begin{comment}
%\vspace{-0.05cm}
\begin{itemize}
\item To the best of our knowledge, it is the first time to directly tackle the challenge induced from varying visage-styles of different speakers, in video-to-speech synthesis by separating the identity and speech content.
\end{itemize} 
%\vspace{-0.45cm}
%Focusing independently on modeling the speech content and the visage-style, the proposed network becomes more robust to predict speech from silent talking face videos of unseen and multiple speakers.
\begin{itemize}
\item We design a speech-visage feature selection for masking identity attributes from a talking face video while maintaining speech content, and vice versa. 
\end{itemize} 
%\vspace{-0.45cm}
%We also apply the module in a multi-view fashion so that the different factors (\eg, age, gender, and etc.) of identity and speech content can be considered during the feature selection. 
\begin{itemize}
\item To guarantee the disentanglement of speech content from identity, we propose two learning methods: visual-identification and audio-identification. %The visualization results of learned representations verify the effectiveness of the proposed framework on disentangling the two factors.
\end{itemize}  %
\end{comment}

%\vspace{-0.4cm}
\section{Related Work}
%\vspace{-0.1cm}
\subsubsection{Video to Speech Synthesis.}
Speech synthesis from silent talking faces is one of the lip-reading techniques that have been consistently studied \cite{michelsanti2020vocoderbased,yadav2021speechpredictioninsilent}. The initial approach \cite{ephrat2017vid2speech} presented an end-to-end CNN-based model that predicts the speech audio signal from a silent talking face video and significantly improved the performance than the methods using hand-crafted visual features \cite{milner2015reconstructingintell}. Another initial work \cite{ephrat2017improved} proposed reconstructing the speech representation by using both video frames and dense optical flow fields for capturing the dynamics of lip movements. Lip2Audspec \cite{akbari2018lip2audspec} also presented a reconstruction-based video-to-speech synthesis method with autoencoders. 1D GAN-based methods \cite{vougioukas2019ganbased,mira2021miragan} were proposed to directly synthesize a raw waveform from the lip movements video. Lip2Wav \cite{prajwal2020lip2wav} introduced a well-known sequence-to-sequence architecture into video-to-speech synthesis to capture the context. Memory \cite{kim2021memory,hong2021speech} proposed to use a multi-modal memory network to associate audio modalities during the inference. Distinct from the previous methods, we try to disentangle the identity characteristics and speech content from a silent talking face video for video-to-speech synthesis. %By adding the visage-style (identity characteristics) on the extracted content, the proposed method can synthesize speech with diverse visage-styles, and of high intelligibility even if a video of an unseen speaker is given.

%\vspace{-0.5cm}
\subsubsection{Feature Selection.}
Feature selection has become a focus of many research areas that utilize huge amounts of high-dimensional data. Early works \cite{li2017featureselection,guyon2003introductiontovariable} initially surveyed feature selection and extraction techniques for improving learning performance, increasing computational efficiency, decreasing memory storage, and building better generalized models. %They roughly categorized feature selection into three methods: wrapper, filter, and embedded methods. 
Among a number of different techniques, deep learning-based feature selection methods are hybrid feature selection methods that ensemble different feature selection algorithms to construct a group of feature subsets \cite{li2017featureselection}. Deep feature selection \cite{li2016deepfeatureselection} selected features by imposing a sparse regularization term to select nonzero weights features at the input level. Another work \cite{roy2015featureselectionusingDNN} proposed a method to assess which features are more likely to contribute to the classification phase. In recent research for feature selection, attention-based feature selection \cite{gui2019afs} was proposed to build the correlation that best describes the degree of relevance of the target and features. %Their feature selection module is composed of an attention module and a learning module, which together build the correlation that best describes the degree of relevance of the target and features.
Most recently, a feature mask module \cite{liao2021featureselectionusingbatchwiseattentuation} is proposed that considers the relationships between the original features by applying a feature mask normalization. %Moreover, a batch-wise attenuation forces the same feature mask for all samples within a training batch during each iteration, which aims better feature selection than the conventional sample-wise attention mechanism.
By adopting the feature selection concept, this paper attempts to select identity-relevant and content-relevant features from the visual representations. %We verify the effectiveness of the proposed feature selection method by visualizing that the selected identity features are distinct from each identity.

\begin{figure*}[t]
	\begin{minipage}[b]{1.0\linewidth}
		\centering
		\centerline{\includegraphics[width=11.5cm]{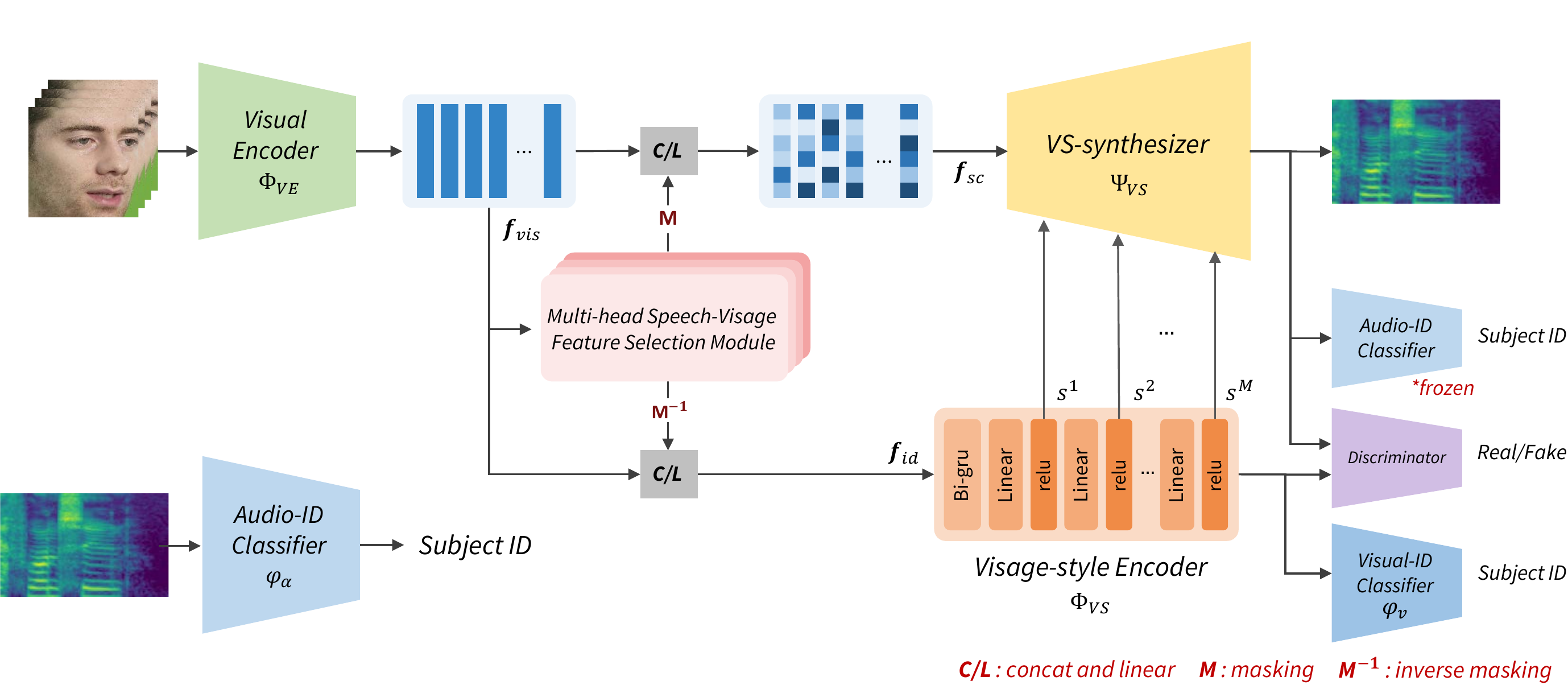}}
	\end{minipage}
	\caption{Overall architecture of the proposed method, containing multi-head speech-visage feature selection and visage-style synthesizer}
	\label{fig:1}
\end{figure*}
%##################################################
%\vspace{-0.1cm}
\section{Proposed Method}
%\vspace{-0.2cm}
Suppose we are given a sequence of silent talking face video $\bm{x}\in\mathbb{R}^{T\times H\times W\times 3}$ with length $T$, height $H$, and width $W$. The goal of our work is to reconstruct a mel-spectrogram $\bm{y}\in\mathbb{R}^{F\times S}$ that matches the input silent talking face frames, where $F$ and $S$ represents the spectral dimension of the mel-spectrogram and the frame length, respectively. The main objective of our learning problem is to disentangle the speech content and the visage-style (\ie, identity) from a silent talking face video, and to synthesize speech by jointly incorporating the two disentangled representations. Hence, it is for enhancing the robustness of the model to unseen speakers and bringing the advantage of generating speech of different visage-styles with fixed speech content. \cref{fig:1} shows the overview of the proposed framework. It contains two major modules: multi-head speech-visage feature selection and visage-style based synthesizer.

%\vspace{-0.4cm}
\subsection{Speech-visage feature selection}
%\vspace{-0.1cm}
When a silent talking face video is given, humans discriminate the entire appearances of the speaker that represent the speaker’s character (\eg, gender and age) and the lip movements, to associate the speech. 
%Then, they predict the speaker’s voice from the appearance and associate the speech by watching the lip movements. 
Motivated from the human cognitive system \cite{chen2001audiovisual}, speech-visage feature selection module is designed to discriminate between human lip movements and visage-styles.

To this end, a visual encoder $\Phi_{VE}$ firstly extracts visual feature $\bm{f}_{vis}$ from a silent talking face video $\bm{x}$ with the dimension of embedding $C$,
\begin{align}
    \bm{f}_{vis}=\Phi_{VE}(\bm{x}) \in\mathbb{R}^{T\times C}.
\end{align}

From $\bm{f}_{vis}$, the proposed speech-visage feature selection module chooses the speech content information while leaving out the identity information, by producing a speech selective mask $\bm{\bar{w}}$. 
Inspired by the modern deep feature selection method \cite{gui2019afs}, the speech selective mask $\bm{\bar{w}}$ is produced with two steps, non-linear transformation and normalization.
Firstly, a non-linear transformation, $\phi_{trans}$ (\ie  LSTM), is applied to the visual features $\bm{f}_{vis}$ to capture the importance of each feature having on speech content,
\begin{align}
    \bm{w}=\phi_{trans}(\bm{f}_{vis}) \in\mathbb{R}^{T\times C}.
\end{align}
%where $\bm{w}\in\mathbb{R}^{T\times C}$. %We use an LSTM for the non-linear transformation to capture the dynamics of lip movements.

%------------------------------------ Figure 2
%#################################################
\begin{figure}[t]
	\begin{minipage}[b]{1.0\linewidth}
		\centering
		\centerline{\includegraphics[width=7.3cm]{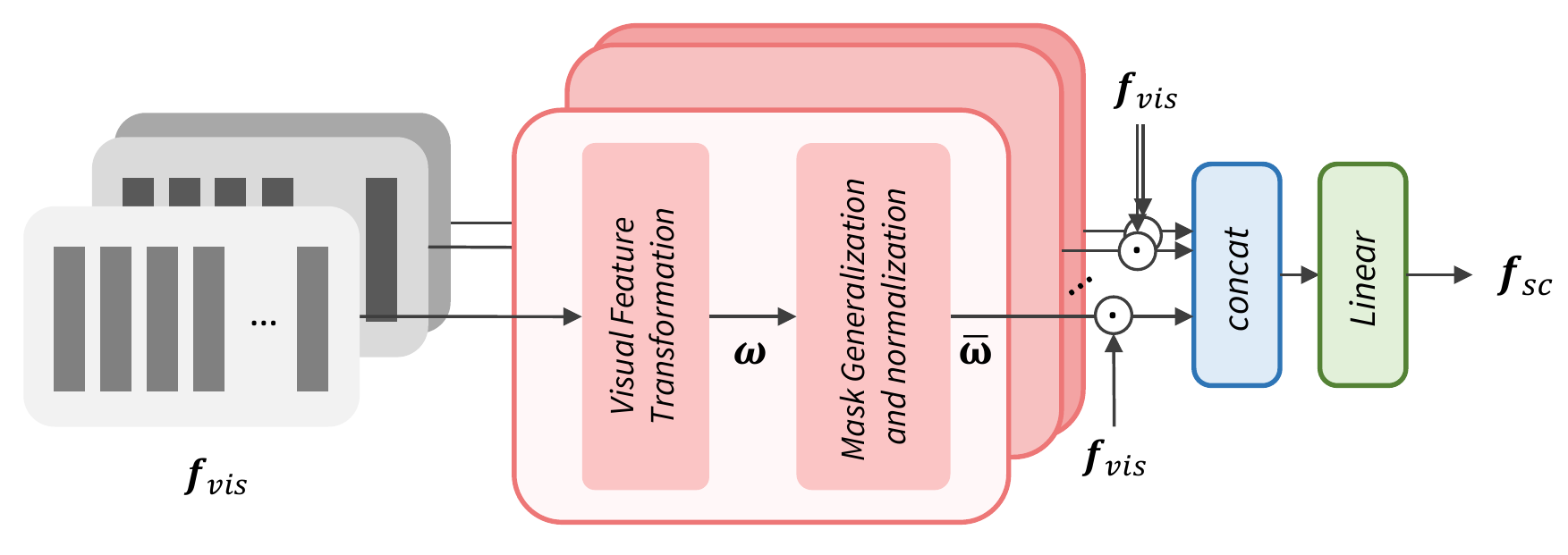}}
	\end{minipage}
	\caption{Multi-head speech-visage feature selection module}
	\label{fig:2}
\end{figure}
%##################################################

Next, mask generalization and normalization are performed to prevent the speech selective mask from being biased to the batch-wise input visual features during training \cite{liao2021featureselectionusingbatchwiseattentuation}. This enables extracting the generalized vector from all samples. For the normalization, the softmax function is utilized to extract the importance score of the speech content of $\bm{f}_{vis}$,
\begin{align}
    \bm{\bar{w}}=Softmax(\frac{1}{B}\sum_{i=1}^B\bm{w_i}),
\end{align}
where $\bm{w_i}$ represents the transformed visual feature of $i$-th sample in the mini-batch size of $B$. Note that the generalization on mini-batch is performed during training only. Then, the speech selective mask is applied to the embedded visual feature $\bm{f}_{vis}$ to select the speech content feature as follows,
\begin{align}
    \bm{f}_{sc}=\phi_{sc}(\bm{\bar{w}}\odot\bm{f}_{vis}) \in\mathbb{R}^{T\times C},
\end{align}
where the $\phi_{sc}$ is an embedding layer, $\odot$ represents element-wise multiplication, and $\bm{f}_{sc}$ represents the selected speech content feature. Since the speech selective mask $\bm{\bar{w}}$ only attends to the speech content information, making the mask opposite, $\bm{\bar{w}}_c=1-{\bm{\bar{w}}}$, can produce the opposite of the speech content, namely the identity feature $\bm{f}_{id}$:
\begin{align}
    \bm{f}_{id}=\phi_{id}(\bm{\bar{w}}_c\odot\bm{f}_{vis}) \in\mathbb{R}^{T\times C},
\end{align}
where $\phi_{id}$ represents a linear layer that embeds the selected identity feature.

%\vspace{-0.4cm}
\subsubsection{Multi-head speech-visage feature selection.}
Due to the multiple characteristics, such as gender and age, of a speaker in regard to the identity and speech content, viewing multiple aspects of the visual face features can enable a better selection of both the speech content and identity. To enhance the feature selection procedure, the speech-visage feature selection can be employed in a multi-view fashion that produces $N$ different speech selective masks, $\{\bm{\bar{w}}^1, \dots, \bm{\bar{w}}^N\}$, as shown in \cref{fig:2}. Similar to the multi-head attention \cite{vaswani2017attention}, our multi-view design allows the model to jointly consider the information with different aspects (\eg, gender and age). The multi-view speech-visage feature selection procedure can be written as,
\begin{align}
    \bm{f}_{sc}=\phi_{sc}([\bm{\bar{w}}^1\odot\bm{f}_{vis}, \dots, \bm{\bar{w}}^N\odot\bm{f}_{vis}]),
\end{align} 
where $[\,,]$ represents concatenation in the channel dimension.
Similarly, we also utilize the inverse of multi-view speech selective masks to obtain the identity features,
%\vspace{-0.1cm}
\begin{align}
    \bm{f}_{id}=\phi_{id}([\bm{\bar{w}}_c^{1}\odot\bm{f}_{vis}, \dots, \bm{\bar{w}}_c^{N}\odot\bm{f}_{vis}]).
\end{align}
We investigate the effect of using multiple speech selective masks in Section 4.3.

%------------------------------------ Figure 1
%#################################################
\begin{figure*}[t]
	\begin{minipage}[b]{1.0\linewidth}
		\centering
		\centerline{\includegraphics[width=12cm]{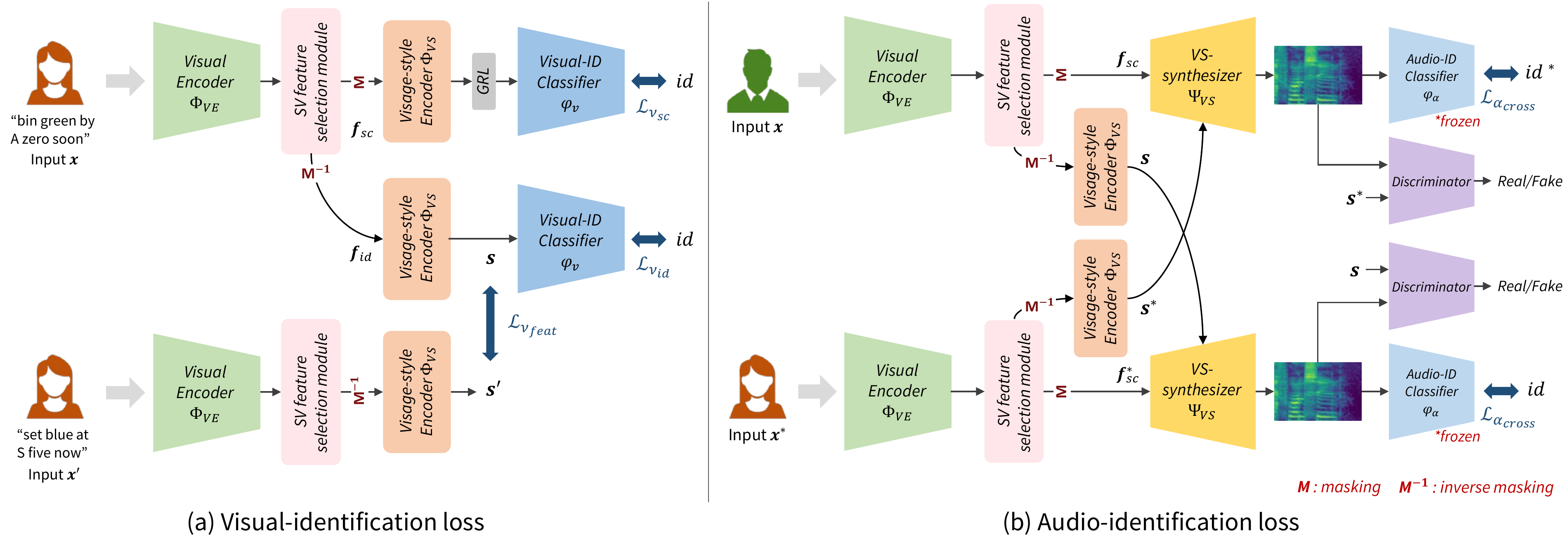}}
	\end{minipage}
	\caption{Visualization of (a) visual-identification loss and (b) audio-identification loss}
	\label{fig:3}
\end{figure*}
%##################################################

%\vspace{-0.2cm}
\subsection{Visage-style based synthesizer}
The speech content features $\bm{f}_{sc}$ contain the correct words of speech and the identity features $\bm{f}_{id}$ have the visage-style of a certain speaker. During generation, the speech content should be maintained and only the style should be coated. Therefore, our generation objective is similar to that of style transfer \cite{gatys2016image,huang2018MUNIT,huang2017adain}. For this purpose, we employ a style-based generator, namely Visage Style-based synthesizer (VS-synthesizer), which reconstructs the mel-spectrogram with respect to the speech content features $\bm{f}_{sc}$ clothed in the encoded identity features $\bm{f}_{id}$. 
For the style encoder, a visage-style encoder $\Phi_{VS}$ is introduced to sequentially extract visage-style features $\bm{s}=\{s^1, \dots, s^M\}$ from the identity features $\bm{f}_{id}$, where $M$ represents the number of styles which will be embedded into the synthesizer through AdaIN \cite{huang2017adain, karras2019stylegan}. The speech (\ie, mel-spectrogram) $\bm{f}_{mel}\in\mathbb{R}^{F\times S}$ is generated with the following equation,
\begin{align}
    \bm{f}_{mel}=\Psi_{VS}(\bm{f}_{sc}, \bm{s}).
\end{align}

To convert the mel-spectrogram into a waveform, we utilize the Griffin-Lim algorithm \cite{griffin1984griffinlim} which is a well known method for converting linear spectrogram into a waveform. Following \cite{wang2017tacotron}, we use a postnet that learns to convert the mel-spectrogram into a linear spectrogram which is utilized for the Griffin-Lim algorithm. It is trained with the reconstruction loss using ground-truth linear spectrograms.

%\vspace{-0.3cm}
\subsection{Learning to select the speech content}
To guide the proposed speech-visage feature selection module to select the speech content feature while leaving out the identity features, we propose two identification learning methods on different modalities, visual and audio.

%\vspace{-0.4cm}
\subsubsection{Visual-identification learning.}
To guide the visage-style features $\bm{s}$ obtained from the identity features $\bm{f}_{id}$ contain identity-related representation, we apply the identification loss as follows,
%\vspace{-0.1cm}
\begin{align}
    \mathcal{L}_{v_{id}} = CE(\varphi_{v}(\Phi_{VS}(\bm{f}_{id})), id),
\end{align}
% %\vspace{-0.1cm}
where $\varphi_{v}$ is a visual-identity classifier, $CE$ represents the cross-entropy loss, and $id$ is the subject identity. Therefore, both the visage-style and identity features can carry the identity-related information.
In addition to the identification loss, we sample two input talking face videos with the same subject, $\bm{x}$ and $\bm{x}'$. We expect that the two extracted visage-style features from each video, $\bm{s}$ and $\bm{s}'$, to be similar, since the visage-style of the same speaker is not varying. Thus, we apply mean squared error objective function as a feature loss,
% %\vspace{-0.05cm}
\begin{align}
\mathcal{L}_{v_{feat}}=||\bm{s}-\bm{s}'||_2.
\end{align}
%\vspace{-0.3cm}

Finally, to guarantee the disentanglement of speech content and identity representations, the speech content feature $\bm{f}_{sc}$ should not contain the identity representations. To achieve this, we adopt an adversarial learning concept that guides the encoder to learn to deceive a classifier. Specifically, Gradient Reversal Layer (GRL) \cite{ganin2015grl} is added before the visual-identity classifier $\varphi_{v}$ so that the gradient sign is reversed during back-propagation. The loss function of speech content feature can be written as follows, 
%\vspace{-0.1cm}
\begin{align}
    \mathcal{L}_{v_{sc}} = CE(\varphi_{v}(grl(\Phi_{VS}(\bm{f}_{sc}))), id).
\end{align}
%\vspace{-0.1cm}
Therefore, the visual-identity classifier struggles to find the identity information from $\bm{f}_{sc}$ while the speech-visage feature selection module learns to not include the identity information into the speech content features $\bm{f}_{sc}$.
Note that we only utilize the last style (\ie, $s^M$) for the visual-identification learning instead of using all styles to reduce the computational cost. The final visual-identification loss (\cref{fig:3}(a)) is defined as $\mathcal{L}_{v}=\mathcal{L}_{v_{id}}+\mathcal{L}_{v_{feat}}+\mathcal{L}_{v_{sc}}$. 

%\vspace{-0.3cm}
\subsubsection{Audio-identification learning.}
Although we disentangled the identity features $\bm{f}_{id}$ and speech content features $\bm{f}_{sc}$, there is no guidance to properly incorporate the two disentangled representations for generating speech. Therefore, we additionally guide the model with a proposed audio-identification loss at the output side. To this end, a pre-trained audio-identity classifier $\varphi_{a}$ is introduced to recognize the subject of the final synthesized mel-spectrogram,
% %\vspace{-0.05cm}
\begin{align}
    \mathcal{L}_{a_{self}}=CE(\varphi_{a}(\Psi_{VS}(\bm{f}_{sc}, \bm{s})),id).
\end{align}
%\vspace{-0.25cm}

Moreover, we design a cross speech classification learning (\cref{fig:3}(b)); when two input talking face videos with different subjects $\bm{x}$ and $\bm{x}^\ast$ are given, we crossly cloth the visage-style features $\bm{s}$ and $\bm{s}^\ast$ into the speech content features of the different subjects, $\bm{f}_{sc}^\ast$ and $\bm{f}_{sc}$, respectively. Therefore, each generated speech should contain the crossly changed visage-style (\ie, identity). This is guided with the following cross-speech classification loss,
\begin{equation}
    \begin{aligned}
        \mathcal{L}_{a_{cross}}=CE(\varphi_{a}(\Psi_{VS}(\bm{f}_{sc}, \bm{s}^\ast)), id^\ast) \\
        + CE(\varphi_{a}(\Psi_{VS}(\bm{f}_{sc}^\ast, \bm{s})), id)).
    \end{aligned}
\end{equation}
Through the cross-speech classification loss, we can achieve both the disentanglement of speech content and identity and the ability to jointly incorporate the disentangled representations in synthesizing the desired speech. The final audio-identification loss is defined as $\mathcal{L}_{a} = \mathcal{L}_{a_{self}} + \mathcal{L}_{a_{cross}}$.

%\vspace{-0.3cm}
\subsection{Total loss functions}
%\vspace{-0.1cm}
%In addition to the visual- and audio-identification learning methods, a generative adversarial loss and a reconstruction loss are employed to complete our learning.
%\vspace{-0.7cm}
\subsubsection{Adversarial loss.}
We utilize both unconditional and conditional GAN losses \cite{goodfellow2014gan,mirza2014conditionalgan}, where the former makes the generated mel-spectrogram realistic, and the latter guides the mel-spectrogram to match the final visage-style feature, $s^M$,
% %\vspace{-0.1cm}
\begin{align}
    \mathcal{L}_g=logD(\bm{f}_{mel}) + logD(\bm{f}_{mel}, s^M),
\end{align}
and the discriminator loss is defined as,
\begin{align}
\begin{split}
    \mathcal{L}_d&=logD(\bm{y}) + log(1-D(\bm{f}_{mel})) \\ 
    &+ logD(\bm{y}, s^M) + log(1-D(\bm{f}_{mel}, s^M)).
\end{split}
\end{align}
%\vspace{-0.68cm}
\subsubsection{Reconstruction loss.}
Finally, a reconstruction loss is adopted to synthesize the mel-spectrogram containing correct contents. The reconstruction loss is defined as,
\begin{equation}
\begin{aligned}
    \mathcal{L}_{recon}=||\bm{y}-\bm{f}_{mel}||_2+ ||\bm{y}-\bm{f}_{mel}||_1.
\end{aligned}
\end{equation}
%\vspace{-0.68cm}
% %\vspace{-0.1cm}
\subsubsection{Total loss.}
The total loss function for the generator part is the sum of the pre-defined loss functions with the balancing weights $\alpha_1$, $\alpha_2$, $\alpha_3$, and $\alpha_4$, 
\begin{equation}
\begin{aligned}
    \mathcal{L}_{tot}=\alpha_{1}\mathcal{L}_{v} + \alpha_{2}\mathcal{L}_a + \alpha_{3}\mathcal{L}_g + \alpha_{4}\mathcal{L}_{recon}.
\end{aligned}
\end{equation}

%\vspace{-0.5cm}
\section{Experiments}
% %\vspace{-0.2cm}
\subsection{Dataset}
\subsubsection{GRID corpus} \cite{cooke2006grid} dataset is the most commonly used dataset for speech reconstruction tasks \cite{ephrat2017vid2speech,ephrat2017improved,michelsanti2020vocoderbased,akbari2018lip2audspec,vougioukas2019ganbased,mira2021miragan,prajwal2020lip2wav}, containing 33 speakers with 6 words taken from a fixed dictionary. Since we focus on the training with a large number of subjects, we conduct experiments on two different settings: 1) multi-speaker independent (unseen) setting where the speakers in the test dataset are unseen and 2) multi-speaker dependent (seen) setting that all 33 speakers are used all training, validation, and evaluation with 90\%-5\%-5\% split, respectively. For multi-speaker independent setting, we follow the same split as \cite{vougioukas2018endtoendspeechdrivenfacial}.
%\vspace{-0.45cm}
\subsubsection{\bf TCD-TIMIT volunteer} \cite{harte2015tcdtimit} dataset has 59 speakers with about 100 phonetically rich sentences. Similar to the GRID dataset, we use two experimental settings.%: 1) multi-speaker independent setting and 2) multi-speaker dependent setting. 
We utilize the officially provided data split of the TCD TIMIT dataset. Please note that it is the first time to exploit the TCD-TIMIT volunteer dataset in a video-to-speech task, which was not utilized due to its difficulties.
%\vspace{-0.45cm}
\subsubsection{\bf LRW} \cite{chung2016lip} dataset contains up to 1000 utterances of 500 different words, spoken by manifold speakers. Since the original dataset does not provide identity information, we clustered and labeled the speaker information of LRW %using a popular face recognition system of ArcFace \cite{deng2019arcface} trained on MS-Celeb-1M \cite{guo2016ms}. 
Total 17,580 speakers are labeled; train, validation, and evaluation splits are newly generated so that the subjects are completely separated among three splits (20 for test and validation, respectively, and the rest for train). It is also the first time to utilize the identity information with the multi-speaker independent (unseen) splits. The details and splits are available in supplementary materials.

% {\bf Lip2Wav} \cite{prajwal2020lip2wav} dataset contains 5 speakers lecturing 5 different lectures, Chess Analysis, Chemistry Lectures, Deep Learning Lectures, Hardware Security, and Ethical Hacking, each of which lasts about 20 hours. 
%\vspace{-0.2cm}
\subsection{Implementation details}
For both GRID and TCD-TIMIT volunteer datasets, we center-crop \cite{sfd} and resize the video frames to 96$\times$96, and 128$\times$128 for LRW dataset. All of the audio in the dataset are resampled to $16$kHz %and transformed into mel-spectrograms with $80$ mel-filter banks. 
We convert the mel-spectrogram so that the length of the mel-spectrogram is $4$ times longer than that of the video frames. %When converting audio to the mel-spectrogram, we use window size of $800$ and hop size of $160$ for $25$fps videos, and window size of $532$ and hop size of $133$ for $30$fps videos, in order to make the length of the mel-spectrogram $4$ times longer than that of the video frames. 
The architectural details of each module can be found in the supplementary materials.
We use the Adam optimizer \cite{kingma2014adam} with $0.0001$ learning rate, discretely decaying half at step $20000$, $40000$, and $60000$. We choose the number $N$ of multi-head masks to 6 and 9 for multi-speaker independent setting and multi-speaker dependent setting, respectively. The number of styles is set to 3 (\ie, $M=3$). The hyperparameters $\alpha_1$, $\alpha_2$, $\alpha_3$, and $\alpha_4$ are $1.0$, $1.0$, $1.0$, and $50.0$, respectively. For computing, we use a single Titan-RTX GPU.
%------------------------------------ Table 1
%#####################################################
\input{./table/Table1.tex}
%#####################################################
%------------------------------------ Table 2
%#####################################################
\input{./table/Table2.tex}
%#####################################################

% \subsection{Evaluation}
For the evaluation, we use three standard speech quality metrics: Short Time Objective Intelligibility (STOI) \cite{taal2010stoi}, Extended Short Time Objective Intelligibility (ESTOI) \cite{jensen2016estoi} for estimating the intelligibility and Perceptual Evaluation of Speech Quality (PESQ) \cite{pesq}.% We also measure audio-visual synchronization rate of the generated speech with the input lip sequence for the speech content recognition performance comparison. We adopt LSE-D and LSE-C measurements that are distance and confidence score between audio and video features from SyncNet, respectively \cite{prajwal2020lip}. Lastly, 
To verify our generated speech, we conduct a human subjective study through mean opinion scores of naturalness, content accuracy, and voice matching.
%Further, we evaluate our generated speech quality through word error rates (WER) on the experiments on the GRID dataset, using Google Speech-to-Text API to transcribe our generated speech and compare to the real transcripts.
%We adopt Griffin-Lim algorithm \cite{griffin1984griffinlim} to obtain the audio from the spectrogram for fair comparisons with previous works \cite{prajwal2020lip2wav,kim2021memory}.

%\vspace{-0.3cm}
\subsection{Experimental results}
%\vspace{-0.15cm}
\subsubsection{Results in multi-speaker independent setting.}
% quantitative 설명
To verify the robustness of the proposed framework to unseen speakers, we conduct the experiments on a multi-speaker independent setting of the GRID and TCD-TIMIT volunteer datasets, where unseen subjects are utilized for testing. Table \ref{table:1} elaborates the performance comparisons on the GRID dataset. We can clearly see that the proposed method outperforms the state-of-the-art performances.%, attaining 0.567 STOI, 0.308 ESTOI, and 1.373 PESQ. 
For the TCD-TIMIT volunteer dataset, shown in the upper part of Table \ref{table:2}, our proposed method achieved 0.478, 0.217, and 1.410, in STOI, ESTOI, and PESQ, respectively, outperforming the previous works \cite{prajwal2020lip2wav, hong2021speech}.% The quantitative results on the two databases confirm the effectiveness of the proposed method on modeling the speech of unseen speaker.

We additionally conduct a human subjective study through mean opinion scores (MOS) for naturalness, intelligibility, and voice matching.% in order to prove the effectiveness of our proposed framework. 
Naturalness evaluates how natural the synthetic speech is compared to the actual human voice, and intelligibility evaluates how clear words in the synthetic speech sound compared to the actual transcription. For the above two measures, naturalness and intelligibility, we follow the exactly same protocol of the previous works \cite{hong2021speech, prajwal2020lip2wav}. We additionally measure voice matching part that determines how well the results of the proposed model matches the voice of the target speaker.
We use 20 samples obtained from the multi-speaker independent setting of the GRID dataset and ask 16 participants to evaluate 6 different approaches and the ground truth in a 5-point scale. The mean scores with 95\% confidence intervals are shown in Table \ref{table:mos}. Our method achieves the score of 2.96, 3.35, and 3.34 for naturalness, intelligibility, and voice matching, respectively, which are the best among the state-of-the-art methods. Especially, the highest intelligibility means the proposed framework can generate speech containing the right content by disentangling the speech content from the identity representations. Moreover, from the voice matching, we verify that the model can synthesize the proper voices that follow the visages of the subjects even if the subjects are not seen before.
%------------------------------------ Table mos
%#####################################################
\input{./table/Table3.tex}
%#####################################################
%------------------------------------ Table 2
%#####################################################
\input{./table/Table4.tex}
%#####################################################

%\vspace{-0.5cm}
\subsubsection{Results in multi-speaker dependent setting.}
To verify that the effectiveness of the proposed method in a multi-speaker dependent setting, we conduct experiments on the full data of the GRID dataset and the TCD-TIMIT volunteer dataset. Table \ref{table:3} shows the comparison results on the GRID dataset with the previous state-of-the-art method \cite{mira2021miragan}. %The proposed method surpasses the previous method by 0.02, 0.066, and 0.091 on STOI, ESTOI, and PESQ, respectively. 
The results on the TCD-TIMIT volunteer dataset are shown in Table \ref{table:4}. The proposed method achieves the best performances except for ESTOI, but it shows comparable performance with \cite{hong2021speech}. The results in the multi-speaker dependent setting show that the proposed method is effective not only for an unseen speaker but also for multi-speaker.
%------------------------------------ Figure quali
%#################################################
\input{./table/table5.tex}
%#################################################
%------------------------------------ Figure 1
%#################################################
\begin{figure}[t]
	\begin{minipage}[b]{1.0\linewidth}
		\centering
		\centerline{\includegraphics[width=11.5cm]{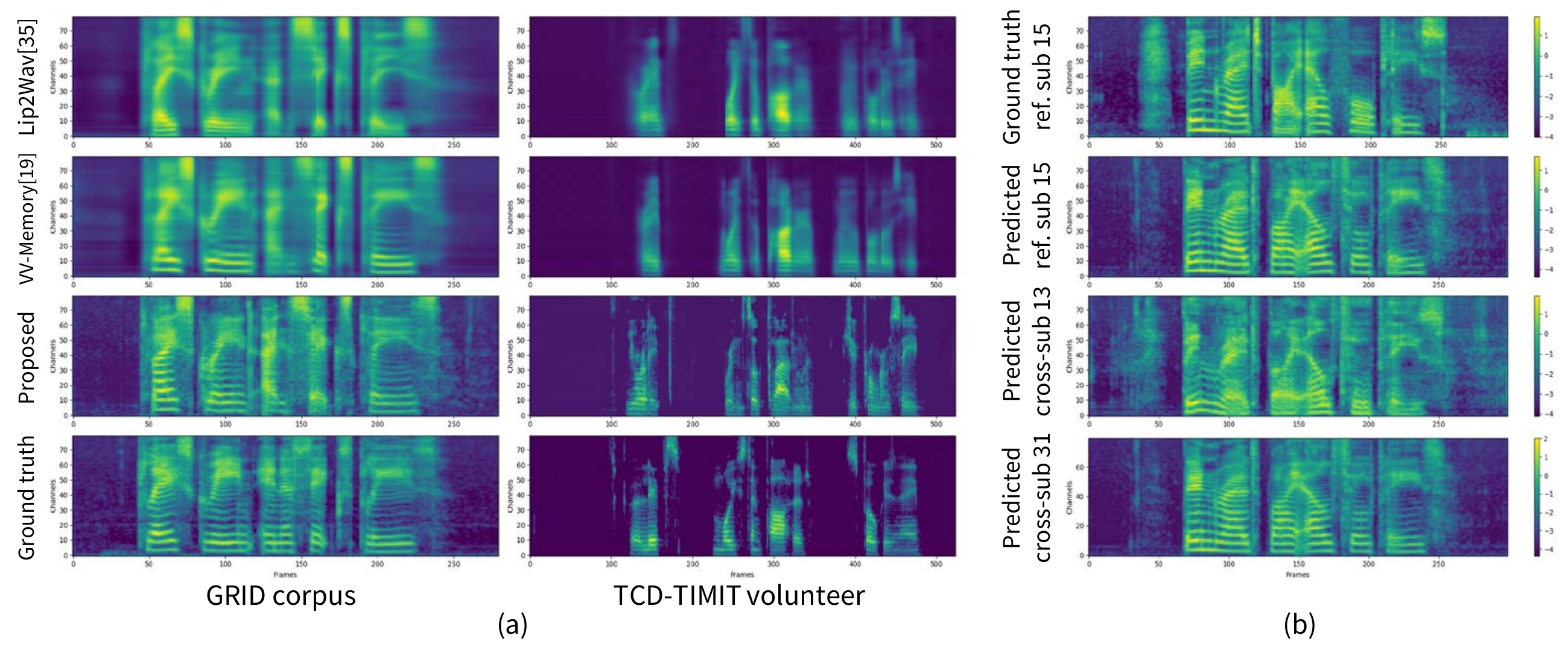}}
	\end{minipage}
	\caption{Qualitative results of (a) generated mel-spectrogram of ground truth, the proposed method, \cite{hong2021speech}, and \cite{prajwal2020lip2wav} in multi-speaker independent setting of GRID corpus and TCD-TIMIT datasets and (b) the ground truth and the generated mel-spectrogram by changing the reference speaking-style features of subject id 15 (female) with that of subject id 13 (male), and that of subject id 31 (female)}
	\label{fig:5}
\end{figure}
%##################################################
%\vspace{-0.5cm}
\subsubsection{Results on dataset with a large number of subjects.}
We additionally conduct an experiment on LRW dataset which contains 17,580 subjects to verify the generalization of the proposed model to new large unseen speakers. Table \ref{table:lrw} shows the performance in multi-speaker independent (unseen) setting on LRW. This even indicates the comparable performance to the results reported in Lip2Wav \cite{prajwal2020lip2wav} (0.543 STOI, 0.344 ESTOI, and 1.197 PESQ) which has performed the experiments on LRW dataset with the original seen setting that contain overlapped subjects in all train, validation, and test splits. This proves that our model works well on dataset with a very large number of subjects with diverse vocabulary, thus generalizing our model's performance. The audio samples of the generated speech of LRW are available in supplementary materials. %The additional comparison results in unseen setting are reported in supplementary materials.

%Since other previous works hardly perform on the unseen dataset setting with hundreds of speakers, we only report our performances.

%\vspace{-0.4cm}
\subsubsection{Qualitative results.}
We visualize the generated mel-spectrogram with the ground truth ones and those from the previous works \cite{hong2021speech, prajwal2020lip2wav}. \cref{fig:5}(a) indicates the generated mel-spectrogram from the multi-speaker independent setting of the GRID and TCD-TIMIT datasets, respectively. 
Additionally, \cref{fig:lrw} shows the generated mel-spectrogram of words \textit{against}, \textit{level}, and \textit{operation} in LRW dataset with the ground truth ones.
It is clearly shown that the generated mel-spectrograms from the proposed method are visually well-matched with the ground truth mel-spectrograms. 

One of our contribution is that we can synthesize speech with different visage-styles by altering the identity features $\bm{f}_{id}$ with others. \cref{fig:5}(b) shows the results of the generated mel-spectrogram with different visage-style features, subject id 13 and 31, which are originally from the subject id 15 of the GRID corpus dataset. When we generate with male speaker's visage-style (\ie, subject id 13) we can observe that the overall frequency of generated mel-spectrogram becomes lower, which means the proposed method can reflect the changed identity features. The audio samples are provided in the supplementary materials.

%------------------------------------ Table 5
%#####################################################
\input{./table/Table6.tex}

%#####################################################
%------------------------------------ Table 6
%#####################################################
\input{./table/Table7.tex}
%#####################################################

%\vspace{-0.5cm}
\subsubsection{Ablation study.}
We analyze the effectiveness of the proposed architecture through ablation studies. We firstly verify two proposed learning methods, visual- and audio-identification, that help to guide the speech-visage feature selection module. Then, we examine that the multi-head speech-visage feature selection technique is more beneficial than the single speech-visage feature selection. Table \ref{table:6} shows the ablation results in the multi-speaker independent setting using the GRID dataset. The baseline is the model that does not apply the speech-visage feature selection, so $\bm{f}_{vis}$ are taken in to both VS-synthesizer and visage-style encoder. %It achieves 0.521, 0.247, and 1.288 on STOI, ESTOI, and PESQ. 
After applying the speech-visage feature selection, the performances increases when both visual- and audio- identification learning methods are adopted. %but with only visual-identification learning, all the performances improve to 0.532 STOI, 0.289 ESTOI, and 1.299 PESQ. The model even attains 0.556, 0.291, and 1.360 on STOI, ESTOI, and PESQ, when both visual- and audio- identification learning methods are adopted. This indicates that learning to disentangle the identity and speech content is beneficial to synthesize the speech of unseen speaker by showing the improved speech quality metrics. 
The highest performances are obtained when multiple selections are adopted with 6 heads in the feature selection. %, the performances increase to 0.567 STOI, 0.308 ESTOI, and 1.373 PESQ. 
The result shows that the multiple masks help the module to discover various attributes of the input visual features, thus yielding better separation of the speech content and identity, which are finally beneficial to reconstruct the speech of diverse speakers.

%\vspace{-0.5cm}
\subsubsection{Effectiveness of multi-heads.}
To analyze the effect of different number of speech selective masks from the multi-head speech-visage feature selection module, we check the performances by differing the number of heads in multi-speaker dependent setting on the GRID dataset, shown in Table \ref{table:7}. While the proposed method with the single speech selective mask achieves the reasonable performance compared to \cite{mira2021miragan} in Table \ref{table:3}, the 9 speech selective masks helps the proposed model attaining the highest performances. %, 0.667 STOI, 0.502 ESTOI, and 1.868 PESQ. 
This means that the sufficient number of the speech selective masks enables our model to separate the speech content and identity.%, thus properly guide the VS-synthesizer to reconstruct the mel-spectrogram.

\begin{figure}[t]
	\begin{minipage}[b]{1.0\linewidth}
		\centering
		\centerline{\includegraphics[width=11cm]{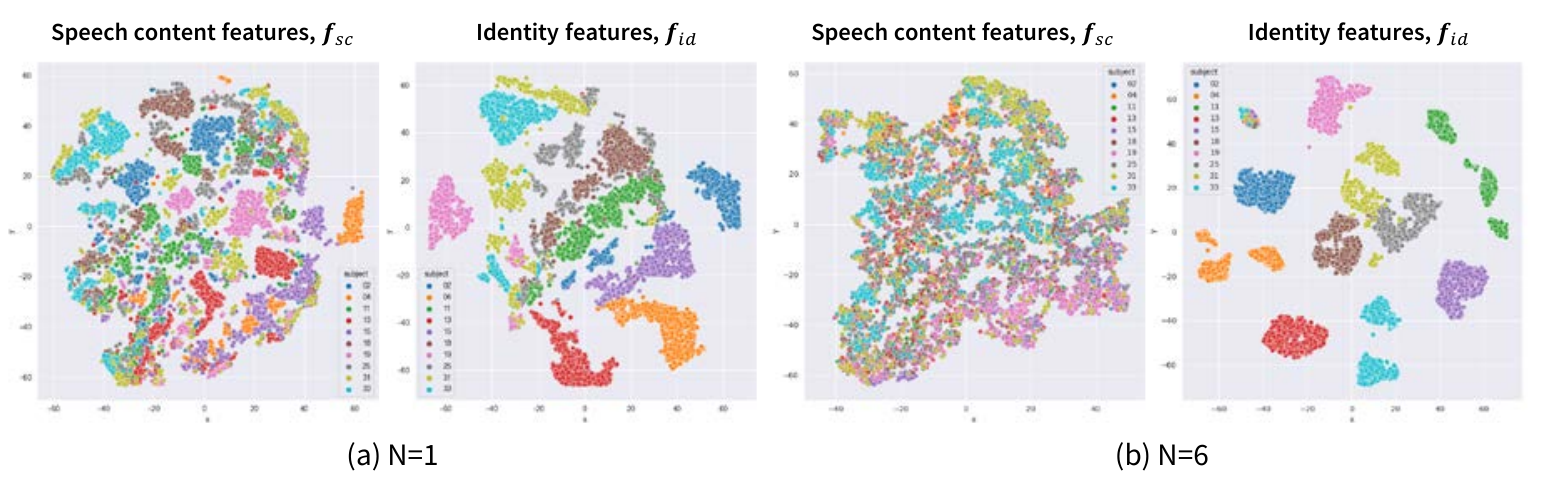}}
	\end{minipage}
	\vspace{-0.8cm}
	\caption{t-SNE \cite{van2008tsne} visualization of speech content features $\bm{f}_{sc}$ and identity features $\bm{f}_{id}$ of (a) single speech visage feature selection procedure (N=1) and (b) multi-head speech visage feature selection procedure (N=6) in regard to the subject ids}
	\label{fig:6}
	\vspace{-0.1cm}
\end{figure}

We additionally visualize the representations of speech content features $\bm{f}_{sc}$ and identity features $\bm{f}_{id}$ in multi-speaker independent setting on the GRID dataset. \cref{fig:6}(a) shows t-SNE \cite{van2008tsne} visualization of two features from the single speech-visage feature selection procedure, $N$=1, and \cref{fig:6}(b) shows the two features from $N$=6.  % where the top two figures represent the features $\bm{f}_{sc}$ and $\bm{f}_{id}$ from the single speech-visage feature selection procedure ($N$=1), and the bottom figures are from the multi-head speech-visage feature selection procedure ($N$=6). 
Each color represents a different subject identity. We can observe that the identity feature $\bm{f}_{id}$ tends to be clustered with the same identity while the speech content feature $\bm{f}_{sc}$ does not, confirming the proposed framework is effective for disentangling the two factors. Moreover, when we increase the number of heads for the speech-visage selection module, the disentanglement is further strengthened as seen in the better-clustered identity features $\bm{f}_{id}$.

%------------------------------------ Figure 1
%#################################################
%##################################################
%------------------------------------ Table 6
%#####################################################
\input{./table/Table8.tex}
%#####################################################
%\vspace{-0.5cm}
\subsubsection{Speaker verification on disentangled features.}
Finally, we perform the speaker verification on the disentangled identity features $\bm{f}_{id}$ and the speech content features $\bm{f}_{sc}$ in multi-speaker independent setting on the GRID. We quantitatively evaluate the content-voice disentanglement quality using the Equal Error Rate (EER) (The lower the EER value, the higher the accuracy) which is commonly used for identity verification. Following \cite{nagrani2017voxceleb}, we find the EER of $\bm{f}_{id}$ to be $29.44\%$ and that of $\bm{f}_{sc}$ to be $33.84\%$ for $N$=1, and $16.90\%$ and that of $\bm{f}_{sc}$ to be $46.48\%$ for $N$=6, shown in Table \ref{table:8}. The results show that the proposed method can well disentangle the identity and speech content representations. With the greater N, the model can disentangle the two features more clearly.

%We observe that the speech content features from the single speech selective mask slightly contain the subject identities. However, through multi-head speech feature selection procedure, the speech content features have almost no subject identity information, and the features in the identity features are well disentangled with respect to the subject identities. This verifies that the multi-head speech-visage feature selection is able to contain multiple aspects of visual features, thus enhancing the selection of both speech content and identity attributes.

% \section{Limitation and Societal Impact}
% We propose a robust video-to-speech synthesis method to an unseen speaker by disentangling the speaker identity and speech content representations. On the other hand, it reveals a limitation of synthesizing speech of ones that are from different datasets. It is probably because of different data distributions of test data from the trained database. This limitation motivates us for further research to solve the data distribution disparity between various datasets. 

% Our work would be vulnerable to the abuse of surveillance applications which lead to a personal privacy invasion. However, it can be utilized much meaningful situations such as corrupted video messages, silent video conferences, and even communication between speech-impaired people.

%\vspace{-0.15cm}
\section{Conclusion}
%\vspace{-0.25cm}
We propose a novel video-to-speech synthesis framework with the speech-visage feature selection, visage-style based synthesizer, and two learning methods. The speech-visage feature selection separates the speech content and speaker identity, and the visage-style based synthesizer utilizes them to adequately reconstruct speech from silent talking face videos. The experimental results on benchmark databases show that the proposed method effectively synthesizes the speech from silent talking face videos of unseen speakers. \cite{selvaraju2017grad}

\medskip
\noindent \textbf{Acknowledgement} This work was supported by the National Research Foundation of Korea (NRF) grant funded by the Korea government (MSIT) (No. NRF-2022R1A2C2005529).

\clearpage
\bibliographystyle{splncs04}
\bibliography{egbib}
\end{document}

%% file: table/table1.tex
\begin{table}[t!]
\caption{Performance comparison in multi-speaker independent setting on GRID}
%\vspace{-0.15cm}
\centering
% \vspace{-0.15cm}
\renewcommand{\arraystretch}{1.1}
\resizebox{0.6\linewidth}{!}{
\begin{tabular}{x{3.3cm}x{1.5cm}x{1.5cm}x{1.5cm}x{1.5cm}x{1.5cm}}
\Xhline{3\arrayrulewidth}
\textbf{Method}          & \textbf{STOI} & \textbf{ESTOI} & \textbf{PESQ} \\ \hline
GAN-based \cite{vougioukas2019ganbased}     & 0.445 & 0.188 & 1.240 \\ % 뺄수도 
Vocoder-based \cite{michelsanti2020vocoderbased}     & 0.537 & 0.227 & 1.230\\
Lip2Wav \cite{prajwal2020lip2wav}       & 0.522 & 0.251 & 1.284\\
VV-Memory \cite{hong2021speech}     & 0.550 & 0.275 & 1.346\\
End-to-end GAN \cite{mira2021miragan}     & 0.553 & 0.269 & 1.372 \\ \hline

\textbf{Proposed model}  &\textbf{0.567} & \textbf{0.308} & \textbf{1.373}
\\ \Xhline{3\arrayrulewidth}
\end{tabular}}
\label{table:1}
\end{table}

% \begin{table}%[]
% \caption{Performance comparison in multi-speaker independent setting on GRID corpus.}
% \vspace{-0.2cm}
% \centering
% % \vspace{-0.15cm}
% \renewcommand{\arraystretch}{1.25}
% \resizebox{0.79\linewidth}{!}{
% \begin{tabular}{x{3.3cm}x{1.4cm}x{1.4cm}x{1.4cm}x{1.6cm}x{1.6cm}}
% \Xhline{3\arrayrulewidth}
% \textbf{Method}          & \textbf{STOI} & \textbf{ESTOI} & \textbf{PESQ}  & \textbf{LSE-D}($\downarrow$) & \textbf{LSE-C}($\uparrow$) \\ \hline
% GAN-based \cite{vougioukas2019ganbased}     & 0.445 & 0.188 & 1.240 & 8.318 & 4.607\\ % 뺄수도 
% Vocoder-based \cite{michelsanti2020vocoderbased}     & 0.537 & 0.227 & 1.230& 7.884& 4.349\\
% Lip2Wav \cite{prajwal2020lip2wav}       & 0.522 & 0.251 & 1.284& 7.103&6.441\\
% Memory \cite{kim2021memory}     & 0.550 & 0.275 & 1.346&7.227&6.146\\
% End-to-end GAN \cite{mira2021miragan}     & 0.553 & 0.269 & 1.372& 7.006 &6.112 \\ \hline

% \textbf{Proposed model}  &\textbf{0.567} & \textbf{0.308} & \textbf{1.373}
% &\textbf{6.955}&\textbf{6.475}\\ \Xhline{3\arrayrulewidth}
% \end{tabular}}
% \label{table:1}
% \end{table}

%% file: table/table2.tex
\begin{table}[t!]
\caption{Performance comparison in multi-speaker independent setting on TCD-TIMIT volunteer dataset}
\centering
%\vspace{-0.15cm}
\renewcommand{\arraystretch}{1.1}
\resizebox{0.59\linewidth}{!}{
\begin{tabular}{x{3.3cm}x{1.5cm}x{1.5cm}x{1.5cm}}
\Xhline{3\arrayrulewidth}
\textbf{Method} & \textbf{STOI}  & \textbf{ESTOI} & \textbf{PESQ} \\ \hline
Lip2Wav \cite{prajwal2020lip2wav} & 0.456 & 0.210 & 1.375 \\
VV-Memory \cite{hong2021speech} & 0.450 & 0.212 & 1.382  \\
\textbf{Proposed model} & \textbf{0.478} & \textbf{0.217} & \textbf{1.410} \\ \hline
\Xhline{3\arrayrulewidth}
\end{tabular}}
\label{table:2}
%\vspace{-0.45cm}
\end{table}

% \begin{table}[t!]
% \caption{Performance comparison in multi-speaker independent setting on TCD-TIMIT volunteer dataset.}
% \centering
% \vspace{-0.15cm}
% \renewcommand{\arraystretch}{1.25}
% \resizebox{0.79\linewidth}{!}{
% \begin{tabular}{x{3.3cm}x{1.4cm}x{1.4cm}x{1.4cm}x{1.6cm}x{1.6cm}}
% \Xhline{3\arrayrulewidth}
% \textbf{Method} & \textbf{STOI}  & \textbf{ESTOI} & \textbf{PESQ}  & \textbf{LSE-D}($\downarrow$) & \textbf{LSE-C}($\uparrow$)\\ \hline
% Lip2Wav \cite{prajwal2020lip2wav} & 0.451 & 0.06 & 1.399&& \\
% Memory \cite{kim2021memory} & 0.450 & 0.212 & 1.382  &&\\
% \textbf{Proposed model} & \textbf{0.478} & \textbf{0.217} & \textbf{1.410} &&\\ \hline
% \Xhline{3\arrayrulewidth}
% \end{tabular}}
% \label{table:2}
% \vspace{-0.3cm}
% \end{table}

%% file: table/table3.tex
\begin{table}[t!]
\centering
\caption{MOS results comparison of the previous methods \cite{vougioukas2019ganbased, michelsanti2020vocoderbased, prajwal2020lip2wav, kim2021memory, mira2021miragan}, the proposed method, and the ground truth}
%\vspace{-0.15cm}
\renewcommand{\arraystretch}{1.1}
\resizebox{0.83\linewidth}{!}{
\begin{tabular}{x{3.3cm}x{2.6cm}x{2.6cm}x{2.6cm}}
\Xhline{3\arrayrulewidth}
\textbf{Method}          & \textbf{Naturalness} & \textbf{Intelligibility} & \textbf{Voice Matching}\\ \hline
GAN-based \cite{vougioukas2019ganbased}     & 1.94$\pm$0.22 & 1.74$\pm$0.21 & 1.37$\pm$0.17  \\
Vocoder-based \cite{michelsanti2020vocoderbased}  & 1.98$\pm$0.16 & 1.68$\pm$0.25 & 1.15$\pm$0.11\\
Lip2Wav \cite{prajwal2020lip2wav}       & 2.71$\pm$0.25 & 2.64$\pm$0.24 & 2.71$\pm$0.23\\
VV-Memory \cite{hong2021speech}     & 2.91$\pm$0.19   & 2.80$\pm$0.23 & 2.85$\pm$0.26  \\
End-to-end GAN \cite{mira2021miragan}     & 2.68$\pm$0.22  & 2.76$\pm$0.26 & 2.18$\pm$0.19\\ 
\textbf{Proposed model}  &\textbf{2.96}$\pm$\textbf{0.28} & \textbf{3.35}$\pm$\textbf{0.34} & \textbf{3.34}$\pm$\textbf{0.27}
\\ \hline
Actual Voice    & 4.28$\pm$0.40  & 4.73$\pm$0.41 & - \\ \Xhline{3\arrayrulewidth}
\end{tabular}}
%\vspace{-0.15cm}
	\label{table:mos}
\end{table}

%% file: table/table4.tex
\begin{table}[t!]
%\vspace{-0.3cm}
\begin{minipage}{.484\linewidth}
\centering
%\vspace{0.21cm}
\caption{Performance comparison in multi-speaker dependent setting on GRID corpus}
%\vspace{-0.05cm}
\renewcommand{\arraystretch}{1.25}
\resizebox{0.999\linewidth}{!}{
\begin{tabular}{x{3.2cm}x{1.2cm}x{1.2cm}x{1.2cm}}
\Xhline{3\arrayrulewidth}
\textbf{Method}          & \textbf{STOI} & \textbf{ESTOI} & \textbf{PESQ} \\ \hline
End-to-end GAN \cite{mira2021miragan}     & 0.647 & 0.436 & 1.777  \\
\textbf{Proposed model}  &\textbf{0.667} & \textbf{0.502} & \textbf{1.868}\\\Xhline{3\arrayrulewidth}
\end{tabular}}
% 	\vspace{-0.1cm}
\label{table:3}
\end{minipage}\hfill
\begin{minipage}{.484\linewidth}
\centering
%\vspace{0.4cm}
\caption{Performance comparison in multi-speaker dependent setting on TCD-TIMIT volunteer dataset}
\renewcommand{\arraystretch}{1.1}
\resizebox{0.999\linewidth}{!}{
\begin{tabular}{x{3.2cm}x{1.2cm}x{1.2cm}x{1.2cm}}
\Xhline{3\arrayrulewidth}
\textbf{Method} & \textbf{STOI}  & \textbf{ESTOI} & \textbf{PESQ} \\ \hline
Lip2Wav \cite{prajwal2020lip2wav} & 0.524 &0.303  &1.545  \\
VV-Memory \cite{hong2021speech}    & 0.555 & \textbf{0.356} & 1.584  \\
\textbf{Proposed model} & \textbf{0.557} & 0.352 & \textbf{1.587}  \\
\Xhline{3\arrayrulewidth}
\end{tabular}}
	\label{table:4}
\end{minipage}
%\vspace{-0.55cm}
\end{table}

%% file: table/table5.tex
\begin{figure}[t]
		\begin{minipage}[c]{0.67\linewidth}
			\centering
			\includegraphics[width=8.2cm]{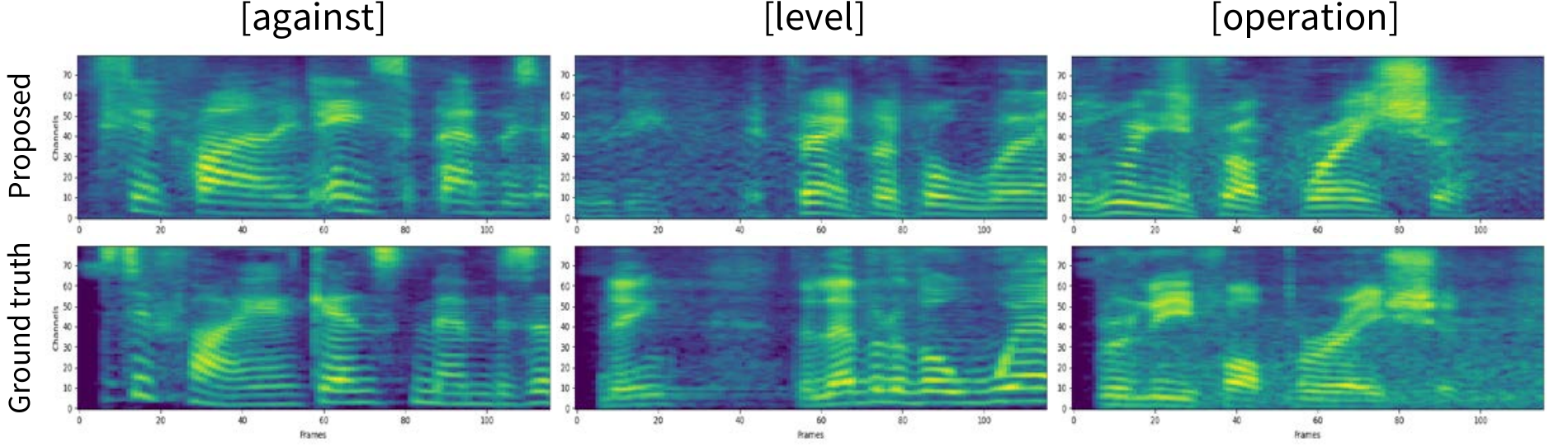}
			\captionof{figure}{Qualitative results of generated mel-spectrogram of ground truth and the proposed method on LRW}
			\label{fig:lrw}
		\end{minipage}
		\hfill
		\begin{minipage}[c]{0.31\textwidth}
			\centering
			\captionof{table}{Performance in multi-speaker independent setting on LRW}
			\renewcommand{\arraystretch}{1.2}
			\resizebox{0.89\linewidth}{!}{
				\begin{tabular}[b]{x{1.85cm}x{1.85cm}}
					\Xhline{3\arrayrulewidth}
					\multicolumn{2}{c}{\textbf{Proposed model}} \\\hline
					\textbf{STOI}  &  0.555  \\ 
					\textbf{ESTOI}&  0.305 \\ 
					\textbf{PESQ}& 1.264 \\ 
					\Xhline{3\arrayrulewidth}
			\end{tabular}}
			\label{table:lrw}
		\end{minipage}
\end{figure}

%% file: table/table6.tex
\begin{table}[t!]
	\renewcommand{\arraystretch}{1.15}
	\renewcommand{\tabcolsep}{2.3mm}
\centering
%\vspace{0.15cm}
	\caption{Ablation study in multi-speaker independent setting on GRID dataset}
%\vspace{-0.15cm}

\resizebox{0.83\linewidth}{!}{
\begin{tabular}{x{1.5cm}x{1cm}x{1cm}x{2cm}x{1.2cm}x{1.2cm}x{1.2cm}}
\Xhline{3\arrayrulewidth}
\multirow{2}{*}[-3.0pt]{\textbf{Baseline}} & \multicolumn{3}{c}{\textbf{Proposed Method}} & & & \\ \cmidrule{2-4}
& \textbf{$\mathcal{L}_v$} & \textbf{$\mathcal{L}_a$} & \textbf{Multi-head} & \textbf{STOI} & \textbf{ESTOI} & \textbf{PESQ}\\ \hline
\cmark & \xmark & \xmark & \xmark & 0.521 & 0.247 & 1.288 \\
\cmark & \cmark & \xmark & \xmark & 0.532 & 0.289 & 1.299 \\
\cmark & \cmark & \cmark & \xmark & 0.556 & 0.291 & 1.360 \\ \hdashline
\cmark & \cmark & \cmark & \cmark & \textbf{0.567} & \textbf{0.308} & \textbf{1.373} \\
\Xhline{3\arrayrulewidth}
\end{tabular}}
% 	\vspace{-0.5cm}
	\label{table:6}
\end{table}

%% file: table/table7.tex
\begin{table}[t!]
	\renewcommand{\arraystretch}{1.15}
	\renewcommand{\tabcolsep}{2.3mm}
\centering
	\caption{Analysis on different number of speech selective masks in multi-speaker dependent setting on GRID dataset}
%\vspace{-0.15cm}
\resizebox{0.62\linewidth}{!}{
\begin{tabular}{x{1.5cm}x{1.2cm}x{1.2cm}x{1.2cm}x{1.2cm}}
\Xhline{3\arrayrulewidth}
\textbf{Metric} & N=1 & N=3 & N=6 & \textbf{N=9}  \\ \hline
\textbf{STOI} &0.651 & 0.648 & 0.653 &\textbf{0.667} \\
\textbf{ESTOI}  &0.489& 0.480 & 0.486 & \textbf{0.502}\\
\textbf{PESQ} & 1.706& 1.738 & 1.767 & \textbf{1.868}\\ %\hdashline
\Xhline{3\arrayrulewidth}
\end{tabular}}
%\vspace{-0.5cm}
	\label{table:7}
\end{table}

%% file: table/table8.tex
\begin{table}[t!]%{r}{5cm}
	\renewcommand{\arraystretch}{1.15}
	\renewcommand{\tabcolsep}{2.3mm}
\centering
% \vspace{-0.1cm}
\caption{The Equal Error Rate (EER) for evaluating the content-voice disentanglement quality}
\vspace{-0.3cm}
\resizebox{0.4\linewidth}{!}{
\begin{tabular}{x{1.6cm}x{1.1cm}x{1.1cm}}
\Xhline{3\arrayrulewidth}
\textbf{EER} (\%) & $\bm{f}_{id}$ & $\bm{f}_{id}$  \\ \hline
N=1 & 29.44 & 33.84  \\
N=6  &16.90& 46.48\\
\Xhline{3\arrayrulewidth}
\end{tabular}}
	\label{table:8}
\end{table}
% 